\pdfoutput=1

\documentclass[11pt]{article}

\usepackage{acl}

\usepackage{times}
\usepackage{latexsym}

\usepackage[T1]{fontenc}

\usepackage[utf8]{inputenc}

\usepackage{microtype}

\usepackage{hyperref}
\usepackage{url}
\usepackage{amsthm}
\usepackage{color}
\usepackage{multirow}
\usepackage{subfig}
\usepackage{graphicx}
\usepackage{algorithm}
\usepackage[noend]{algpseudocode}
\usepackage{cancel}
\usepackage{amsthm}
\usepackage{wrapfig}
\usepackage{booktabs}
\usepackage{makecell}

\usepackage{amsmath}

\usepackage{amssymb}
\usepackage{pifont}
\usepackage{bbm}
\usepackage{setspace}
\usepackage{asymptote}
\usepackage{colortbl} 
\usepackage{soul}     

\newcommand{\ours}{\textsc{Bba}}
\DeclareMathOperator*{\argmax}{arg\,max}

\title{\texorpdfstring{\ours}{}: Bi-Modal Behavioral Alignment for Reasoning with Large Vision-Language Models}

\author{
\textbf{Xueliang Zhao}$^\spadesuit$\thanks{\hspace{1.5mm}Work done during internship at Tencent AI Lab.} \ \ 
\textbf{Xinting Huang}$^\diamondsuit$  \ \ 
\textbf{Tingchen Fu}$^\diamondsuit$ \ \
\textbf{Qintong Li}$^\spadesuit$ \ \
\textbf{Shansan Gong}$^\spadesuit$ \ \
\\
\textbf{Lemao Liu}$^\diamondsuit$ \ \
\textbf{Wei Bi}$^\diamondsuit$  \ \ 
\textbf{Lingpeng Kong}$^\spadesuit$ \\
$^\spadesuit$The University of Hong Kong
\quad
$^\diamondsuit$Tencent AI Lab \\
\texttt{\{xlzhao,lpk\}@cs.hku.hk}\\
}

\begin{document}
\maketitle
\begin{abstract}

Multimodal reasoning stands as a pivotal capability for large vision-language models (LVLMs). The integration with Domain-Specific Languages (DSL), offering precise visual representations, equips these models with the opportunity to execute more accurate reasoning in complex and professional domains. However, the vanilla Chain-of-Thought (CoT) prompting method faces challenges in effectively leveraging the unique strengths of visual and DSL representations, primarily due to their differing reasoning mechanisms. Additionally, it often falls short in addressing critical steps in multi-step reasoning tasks. To mitigate these challenges, we introduce the \underline{B}i-Modal \underline{B}ehavioral \underline{A}lignment (\ours{}) prompting method, designed to maximize the potential of DSL in augmenting complex multi-modal reasoning tasks. This method initiates by guiding LVLMs to create separate reasoning chains for visual and DSL representations. Subsequently, it aligns these chains by addressing any inconsistencies, thus achieving a cohesive integration of behaviors from different modalities. Our experiments demonstrate that \ours{} substantially improves the performance of GPT-4V(ision) on geometry problem solving ($28.34\% \to 34.22\%$), chess positional advantage prediction ($42.08\% \to 46.99\%$) and molecular property prediction ($77.47\% \to 83.52\%$).

\end{abstract}

\begin{figure*}
    \centering
    \begin{minipage}{0.48\linewidth}
        \centering
        \includegraphics[width=\linewidth]{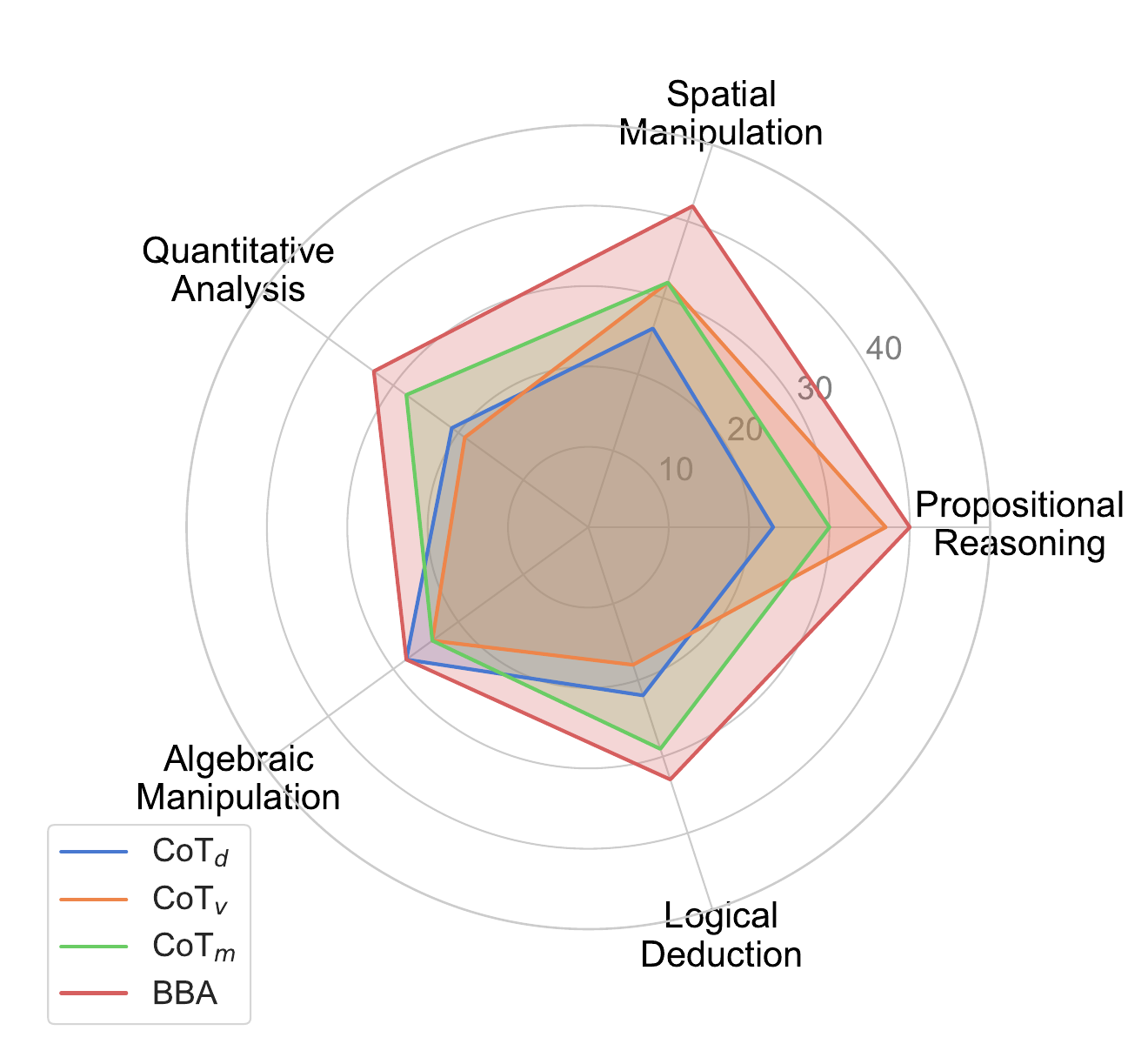}
        \label{fig:left_pilot}
    \end{minipage}\hfill
    \begin{minipage}{0.48\linewidth}
        \centering
        \includegraphics[width=\linewidth]{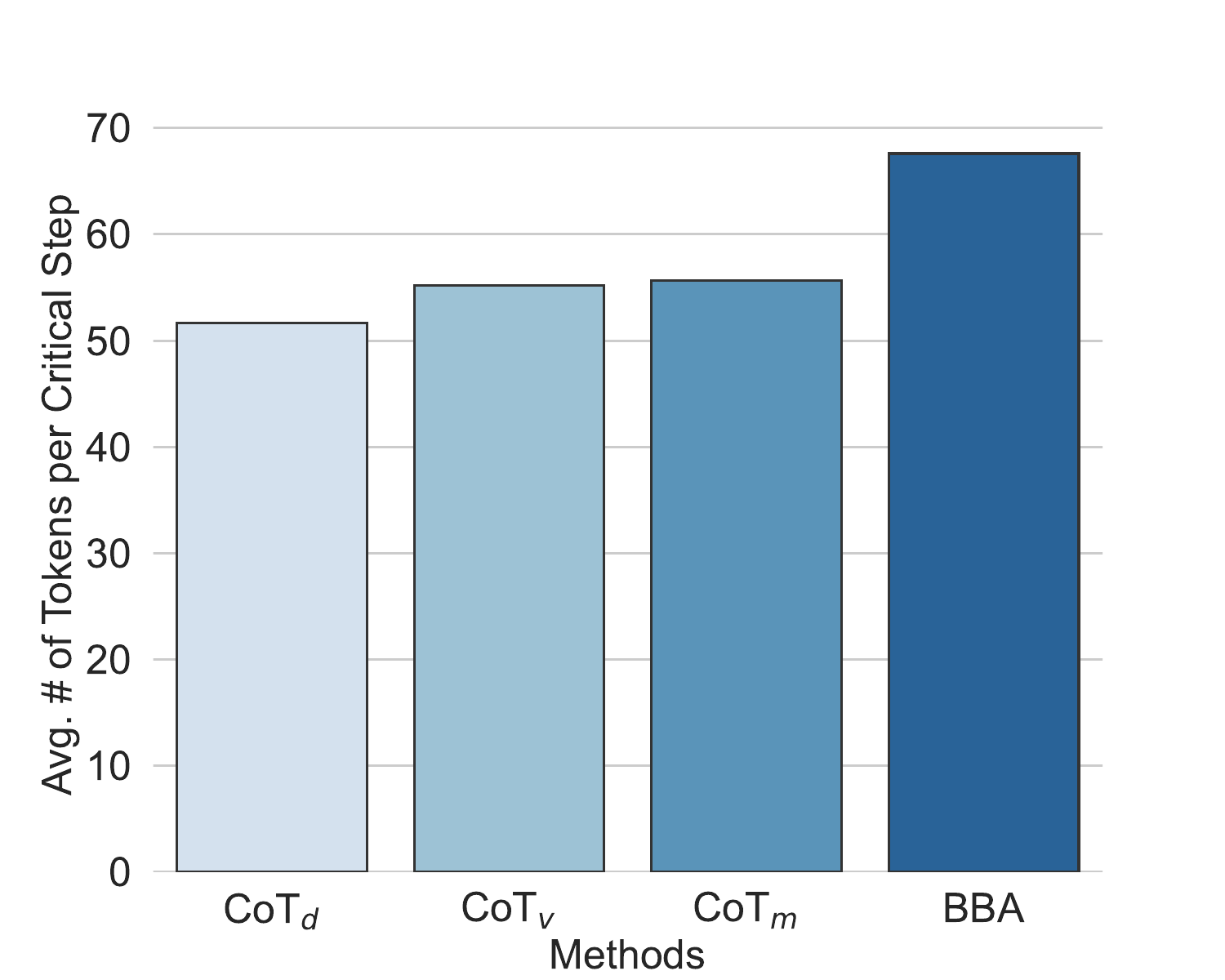}
        \label{fig:right_pilot}
    \end{minipage}
    \vspace{-2mm}
    \caption{Comparative analyses of different methods in problem-solving and critical step detailing. \textbf{Left}: Problem-solving rates across diverse problem types, where CoT$_d$ and CoT$_v$ refer to Chain-of-Thought prompting with DSL and image inputs, respectively, and CoT$_m$ represents the approach combining both inputs. \textbf{Right}: Average number of tokens per critical step across different methods.}
    \vspace{-4mm}
    \label{fig:pilot_studies}
\end{figure*}

\section{Introduction}

The use of domain-specific language (DSL)~\cite{bowman2008asymptote,edwards1994portable,weininger1988smiles} aims to incorporate multimodal information by providing a precise and unequivocal alternative form using text.\footnote{Figure~\ref{fig:pipeline} illustrates an example of a DSL tailored for the geometry domain. Further instances of DSLs can be found in Appendix~\ref{sec:appendix_case}.}
Its application has significantly improved the multimodal reasoning capability, yielding notable improvements in intricate contexts, especially within specialized domains such as symbolic reinforcement learning~\cite{mcgrath2022acquisition,zahavy2023diversifying,ruoss2024grandmaster} and diverse scientific fields~\cite{winter2022smile}. 

Multimodal reasoning is a fundamental capability for large vision-language models (LVLMs)~\cite{openai2023gpt4,yang2023dawn}, crucial for many of their applications. Despite the considerable progress made by LVLMs in multimodal tasks~\cite{lu2023vim,hu2023look}, effectively utilizing them for complex multimodal reasoning, particularly in conjunction with DSLs, remains underexplored. The most direct approach is to feed the LVLMs with both visual data (e.g., images) and its corresponding DSL representation along with the textual queries. They are then guided through the Chain-of-Thought (CoT)~\cite{wei2023chainofthought} prompting to process step-by-step reasoning. However, a significant issue with this approach is that the reasoning processes derived from different modalities are often inconsistent, or even conflicting. This inconsistency limits the ability of LVLMs to effectively integrate the strengths of visual and DSL representations~(\S\ref{sec:pilot_versatile}). Moreover, these models encounter difficulties in executing multi-step reasoning~\cite{wu2023early,liu2023evaluation}, which hampers their effectiveness in addressing critical steps within complex problems~(\S\ref{sec:pilot_critical}).

To address these challenges, we propose a \underline{B}i-Modal \underline{B}ehavioral \underline{A}lignment (\ours{}) prompting method that adeptly integrates DSL into complex multimodal reasoning tasks. \ours{} begins by prompting LVLMs to generate distinct reasoning chains from both visual and DSL representations, and then aligns these chains by resolving inconsistencies, thereby harmonizing the behaviors elicited from various modalities. \ours{} offers two primary advantages. Firstly, it adopts a ``late fusion'' strategy~\cite{ghanem2018activitynet,owens2018audio}, effectively maintaining the inherent strengths of both the direct vision input and the DSL representation. Secondly, \ours{} turns the inconsistency across modalities into a beneficial signal that aids in identifying critical steps within reasoning processes. By revealing where the reasoning chains differ, it efficiently allocates more intermediate tokens to these critical steps by resolving the inconsistencies found.

We evaluate \ours{} on three multimodal reasoning tasks: geometry problem-solving, chess positional advantage prediction, and molecular property prediction. In these diverse applications, \ours{} demonstrated notable relative improvements, with respective performance improvements of 14.26\%, 10.25\%, and 6.30\%.

\section{Pilot Study}
\label{sec:pilot_study}
In this study, we compare three variants of CoT prompting within domains where DSL is available. These variations include: (1) CoT$_v$, which utilizes only images for grounding responses to queries; (2) CoT$_d$, which relies exclusively on DSL representations for grounding; and (3) CoT$_m$, which integrates both images and DSL representations. 
We focus on a selection of mathematical geometry problems from the MATH benchmark~\cite{hendrycks2021measuring}, comprising a total of $187$ problems that incorporate image inputs. 
We then explore the difficulties associated with performing multi-modal reasoning using both images and DSL representations, through an empirical examination of distinct success rates across various problem types and the allocation of tokens for critical reasoning steps.

\subsection{Performance on Fine-grained Types}
\label{sec:pilot_versatile}
Our analysis begins with an assessment of the performance of different models on fine-grained problem types. To this end, we categorically divide the geometry problems based on the primary skills required for resolution, resulting in five categories: (1) Spatial Manipulation, (2) Propositional Reasoning, (3) Logical Deduction, (4) Algebraic Manipulation, and (5) Quantitative Analysis. Additional details on the categorization annotation can be found in Appendix~\ref{sec:appendix_type}. We proceed to calculate and compare the problem-solving rates for each category.

Figure~\ref{fig:pilot_studies} offers a visual comparison of the models' performances across these categories. It is evident that CoT$_v$ and CoT$_d$ exhibit significantly different levels of effectiveness across these problem types. Specifically, CoT$_v$ shows superior performance in tasks involving spatial manipulation and propositional reasoning, while CoT$_d$ excels in logical deduction, algebraic manipulation, and quantitative analysis. This variation in performance can be attributed to the different reasoning mechanisms enabled by each modality. DSL representations provide detailed information (e.g., precise coordinates) that support logic-oriented operations. On the other hand, images provide intuitive visual cues that are more conducive to spatial reasoning tasks. Despite the concurrent use of images and DSL representations, CoT$_m$ does not demonstrate uniform improvements across all problem types, indicating the challenge of aligning reasoning mechanisms across modalities. In \S\ref{sec:method}, we elaborate on \ours{}, which initiates by independently deriving reasoning chains from images and DSL representations, and then aligning these chains by resolving any inconsistencies between them. Unlike CoT$_m$, \ours{} effectively capitalizes on the strengths of both modalities, achieving comprehensive improvements across all identified problem categories.

\subsection{Token Allocation for Critical Steps}
\label{sec:pilot_critical}
In light of recent theoretical advances~\cite{feng2023towards,merrill2023expresssive} indicating the effective allocation of intermediate tokens as pivotal for unlocking the expressive power of models in sequential reasoning tasks, we delve into the allocation of intermediate tokens for addressing critical steps in problem-solving. A critical step in solving mathematical problems is defined as the point at which an essential insight, decision, or application of a method is crucial for obtaining the correct solution, typically involving a significant conceptual leap, strategic theorem application, or key calculation that influences the subsequent problem-solving process. For each problem, we identify all critical steps, categorizing each step in the generated solution as either corresponding to one of the identified critical steps or not, and then sum the tokens for steps within a generated solution that are associated with the same critical step. Details on the annotation of critical steps are provided in Appendix~\ref{sec:appendix_critical}.

Figure~\ref{fig:pilot_studies} demonstrates that merely combining images and DSL representations in inputs is insufficient for effectively allocating more tokens to critical steps, thus reducing the expressive power of LLMs and leading to inferior overall performance (as discussed in \S\ref{sec:exp_main}). We hypothesize that this limitation arises from the current inefficiencies of LLMs in exploring the solution space for complex problems~\cite{yang2023dawn}, resulting in their struggle to accurately identify critical steps. As will be discussed in \S\ref{sec:method_alignment}, \ours{} is more effective in discerning and addressing critical steps by uncovering and reconciling discrepancies among reasoning chains derived from different modalities.

\begin{figure*}[t]
    \centering
    \includegraphics[width=0.9\textwidth]{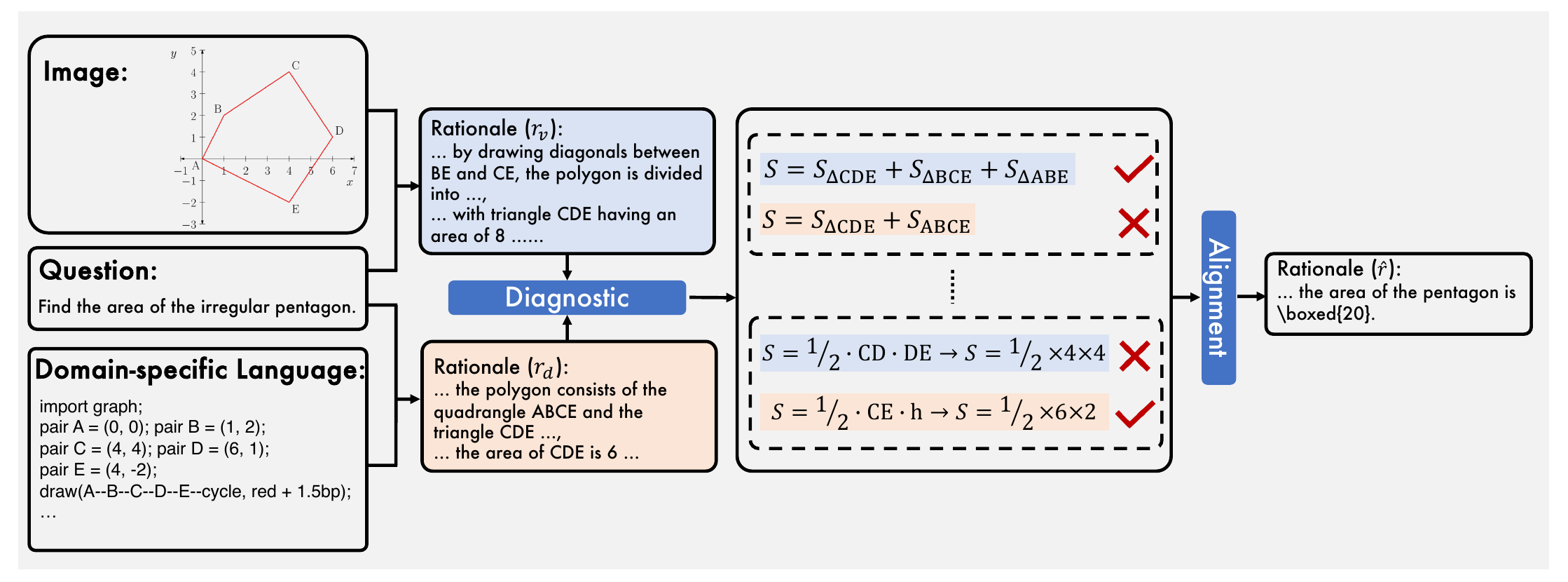}
    \vspace{-2mm}
    \caption{An instantiation of the proposed \ours{} method.}
    \vspace{-4mm}
    \label{fig:pipeline}
\end{figure*}

\section{Preliminaries}
\subsection{Problem Formulation}
\label{sec:formulation}
This study focuses on multi-modal reasoning tasks, specifically where the visual modality is represented as an image, coupled with a DSL that accurately depicts the image. Our objective is to predict an answer to a given question $q$, associated with an image $v$ and a DSL representation $d$, adhering to specific task requirements (e.g., solving mathematical problems). 

The emergence of LVLMs has streamlined this process. Owing to extensive pre-training on trillions of tokens, these models can accurately interpret various instructions and execute the corresponding tasks. In this paradigm, the model parameters are denoted by $\boldsymbol{\theta}$, and the answer $\hat{a}$ is generated as $\hat{a} = \argmax_{a} p(a \mid q, v, d; \boldsymbol{\theta})$, where the inputs are reformulated into well-crafted prompts using specific templates, designed to elicit the desired response from the LVLMs.

\subsection{Chain-of-Thought Prompting}
\label{sec:cot}
Recently, chain-of-thought prompting has gained recognition as an effective technique for enhancing the reasoning capabilities of language models~\cite{wei2023chainofthought}. This method decomposes the original task into two distinct phases: rationale generation and answer prediction. In the rationale generation phase, a rationale $\hat{r}$ is derived as $\hat{r} = \argmax_{r} p(r \mid q,v,d; \boldsymbol{\theta})$, leveraging a query augmented with an instruction designed to initiate stepwise analytical thinking~\cite{kojima2022large}). Subsequently, the answer is often deduced directly from the rationale, utilizing heuristic string-matching methods for precise identification.

\section{Method}
\label{sec:method}
This work aims to tackle two primary challenges in multi-modal reasoning: (1) the integration of the inherent strengths of both visual and DSL representations, and (2) the identification and resolution of critical steps within these tasks.
To address these challenges, we introduce the \ours{} prompting method, an innovative approach that seeks to unleash the power of DSL in enhancing complex multi-modal reasoning tasks.  Figure~\ref{fig:pipeline} offers an overview of our proposed methodology. \ours{} initiates by employing LVLMs to generate reasoning chains separately from visual and DSL inputs. Subsequently, these chains proceed through an alignment phase, wherein inconsistencies are identified and reconciled, ensuring the harmonization of behaviors derived from each modality.

\paragraph{Road Map.} The rest of this section is structured as follows: We begin by detailing the process of eliciting reasoning chains from both vision and DSL representations in \S\ref{sec:method_eliciting}. This is followed by an elaboration on diagnosing and rectifying inconsistencies between these reasoning chains and the methods of aligning behaviors from different modalities in \S\ref{sec:method_alignment}. 
Lastly, in \S\ref{sec:method_insight}, we detail how \ours{} effectively identifies and addresses critical steps in the reasoning process.

\subsection{Bi-Modal Behavior Eliciting}
\label{sec:method_eliciting}

The objective of this phase is to effectively harness the unique strengths of vision and DSL representations in answering a given question. Unlike vanilla CoT prompting, which intermingles the reasoning processes of these two modalities, \ours{} seeks to elicit reasoning chains from each modality independently. This approach allows the vision-based reasoning chain to deliver more credible steps in intuitive and spatial reasoning, while the DSL-based reasoning chain provides steps with greater reliability in precise computation. The formal definition of this process is as follows:
\begin{equation}
\begin{aligned}
r_{\text{v}} &= \argmax_{r} p(r \mid q,v; \boldsymbol{\theta}) \\
r_{\text{d}} &= \argmax_{r} p(r \mid q,d; \boldsymbol{\theta}).
\end{aligned}
\end{equation}
where $r_{\text{v}}$ and $r_{\text{d}}$ represent the reasoning chains derived from the vision and DSL representations, respectively.

\subsection{Behavior Alignment}
\label{sec:method_alignment}
This phase is centered on aligning the reasoning chains from different modalities to capitalize on the best of both worlds in multi-modal reasoning. 
We initiate this process with diagnostic checks to uncover inconsistencies between the chains, including variances in intermediate steps and the final answers. Following this, an aligned reasoning chain is created by addressing the discrepancies identified in the diagnostics. When different methods produce conflicting results, it often indicates an error in at least one approach. The divergence point then becomes a crucial indicator of where deeper understanding or more meticulous application of principles is necessary. The model is subsequently instructed to thoroughly examine the derivations from both modalities and ascertain accurate conclusions. The diagnostic results are formally obtained as follows:
\begin{equation}
r_{\text{inc}} =\argmax_r p(r \mid r_{\text{v}}, r_{\text{d}};\boldsymbol{\theta}),
\end{equation}
where $r_{\text{inc}}$ denotes the rationale for inconsistencies identified during the diagnostic process.
Next, the formation of the aligned reasoning chain is defined as:
\begin{equation}
\hat{r} =\argmax_r p(r \mid r_{\text{v}}, r_{\text{d}}, r_{\text{inc}}; \boldsymbol{\theta})
\end{equation}
where the final rationale $\hat{r}$ includes the definitive answer $a$ within special tokens.

\subsection{Discussion}
\label{sec:method_insight}

The strengths of \ours{} can be mainly attributed to its capability to address critical steps in multi-step reasoning problems. \ours{} excels in addressing critical steps primarily due to two reasons: (1) the critical step is more easily identified by contrasting different solutions, revealing their divergences; and (2) learning from these differences allows for a more efficient allocation of intermediate tokens to these critical steps. Drawing from cognitive learning principles observed in humans, it is a plausible extrapolation that identifying and rectifying disparities between various methods fosters a deeper comprehension of essential aspects of a problem~\cite{munzar2021elementary}. Furthermore, encountering and acknowledging mistakes enhances the reasoning process, paralleling human problem-solving strategies. This not only deepens the understanding but also facilitates the allocation of additional reasoning tokens, thereby amplifying the model's capacity to resolve critical steps~\cite{feng2023towards,merrill2023expresssive}.

\begin{table*}
\centering
\resizebox{0.9\linewidth}{!}{
\begin{tabular}{lcccccc}
\toprule
Methods & With DSL  & With Figure & \textbf{G-MATH} & \textbf{ChessAdv} & \textbf{MUTAG} & Avg. \\ \midrule
CoT$_v$~\cite{wei2023chainofthought} & \ding{55}  & \ding{51} & 23.53 & 40.98 & 75.82 & 46.56 \\
CoT$_d$~\cite{wei2023chainofthought} & \ding{51}  & \ding{55} & 23.12 & 38.80 & 76.92 & 46.01 \\
Plan-and-Solve~\cite{wang2023plan} & \ding{51} & \ding{55}  & 25.67 & 42.62  & 78.57 & 48.73 \\
Least-to-Most~\cite{zhou2022least} & \ding{51} & \ding{55} & 25.13  & 38.25 & 73.63 & 45.47 \\ \midrule
CoT$_m$~\cite{wei2023chainofthought} & \ding{51} & \ding{51} & 28.34  & 42.08 & 77.47 & 49.09 \\
CCoT~\cite{mitra2023compositional} & \ding{51} & \ding{51} & 26.74 & 39.34 & 68.68 & 44.75 \\
DDCoT~\cite{zheng2023ddcot} & \ding{51} & \ding{51} & 29.95  & 37.70 & 73.08 & 46.74 \\ \midrule
\ours{}~(Ours) & \ding{51} & \ding{51} & \textbf{34.22} & \textbf{46.99} & \textbf{83.52}  & \textbf{54.71} \\
\bottomrule
\end{tabular}
}
\caption{Evaluation results for geometry problem-solving (\textbf{G-MATH}), chess positional advantage prediction (\textbf{ChessAdv}), and molecular property prediction (\textbf{MUTAG}), including average performance. Numbers in bold denote the best performance.} 
\label{tab:exp_main}
\end{table*}

\begin{table*}
\centering
\resizebox{0.8\linewidth}{!}{
\begin{tabular}{lcccccc}
\toprule
Methods & With DSL  & With Figure & \textbf{G-MATH} & \textbf{ChessAdv} & \textbf{MUTAG} & Avg. \\ \midrule
\ours{}~(Ours) & \ding{51}  & \ding{51} & \textbf{34.22} & \textbf{46.99} & \textbf{83.52} & \textbf{54.71} \\ \midrule
-diagnostic & \ding{51} & \ding{51} & 32.09  & 41.53 & 78.57 & 50.54 \\ 
-visual & \ding{51} & \ding{55}  & 28.34 & 37.70 & 61.54 & 42.39 \\
-dsl & \ding{55} & \ding{51} & 27.27 & 36.07 & 75.82 & 46.20 \\
\bottomrule
\end{tabular}
}
\caption{Ablation study results with best performances highlighted in bold.}
\label{tab:exp_abl}
\end{table*}

\section{Experiments}
\subsection{Datasets and Evaluation}

We assess the efficacy of \ours{} across three multi-modal reasoning tasks spanning distinct domains: geometry problem-solving, chess positional advantage prediction, and molecular property prediction.

\paragraph{Geometry Problem-Solving.}

This task involves predicting a free-form solution to a given geometry problem. We utilize the geometry subset of the MATH benchmark~\cite{hendrycks2021measuring} for this task, selecting only those problems that include Asymptote code~\cite{bowman2008asymptote}, a domain-specific language (DSL) used for depicting geometric figures. This process resulted in a dataset of $187$ problems, which we refer to as \textbf{G-MATH}. The official evaluation script from the MATH benchmark is employed to compute accuracy by comparing the predicted answers with the correct answers.

\paragraph{Chess Positional Advantage Prediction.}
The objective in chess positional advantage prediction is to classify a given chessboard state as being advantageous for White, advantageous for Black, or balanced. This task evaluates the model's capacity to correlate with the actual value of a chessboard state, determined by chess engines after extensive analysis. For evaluation, we compiled a dataset of $183$ game snippets, applying Stockfish 15 at a search depth of 18 to assess the winning probability for the white pieces. We classified the winning probabilities into three intervals: 0–33\% indicating an advantage for Black, 34–66\% denoting a balanced state, and 67–100\% suggesting an advantage for White. We refer to this dataset as \textbf{ChessAdv}, employing Forsyth-Edwards Notation~(FEN)~\cite{edwards1994portable} as the DSL for this domain. Classification accuracy serves as the evaluation metric.

\paragraph{Molecular Property Prediction.}
Molecular property prediction focuses on determining whether a molecule exhibits a certain property based on its molecular graph. The \textbf{MUTAG} benchmark dataset~\cite{debnath1991structure} is used for this purpose, comprising $188$ chemical compounds categorized into two classes based on their mutagenic effects on a bacterium. The Simplified Molecular-Input Line-Entry System~(SMILES)~\cite{weininger1988smiles} is utilized as the DSL in this domain, with classification accuracy as the metric for evaluation.

\subsection{Baselines}
For comparative evaluation, we adopt the following baselines:
\paragraph{DSL or Visual-Only Methods.} (1) \textbf{CoT$_v$}: Implements chain-of-thought prompting~\cite{wei2023chainofthought}, omitting DSL representations and relying solely on images; (2) \textbf{CoT$_d$}: Utilizes chain-of-thought prompting, excluding images to focus exclusively on DSL representations; (3) \textbf{Plan-and-Solve}: Formulates a plan to segment the overall task into manageable subtasks for sequential execution~\cite{wang2023plan}; and (4) \textbf{Least-to-Most}: Breaks complex problems into simpler, sequential subproblems, leveraging solutions of preceding subproblems to facilitate solving subsequent ones~\cite{zhou2022least}. 

\paragraph{Integrated DSL and Visual Methods.} (1) \textbf{CoT$_m$}: Employs chain-of-thought prompting using a combination of both DSL representations and images; (2) \textbf{CCoT}: enhances compositional reasoning by integrating visual and DSL inputs, substituting the scene graph with DSL for fair comparison~\cite{mitra2023compositional}; (3) \textbf{DDCoT}: Introduces negative-space prompting and multimodal reasoning by dividing cognitive tasks between reasoning and recognition, enhancing reasoning with visual recognition capabilities~\cite{zheng2023ddcot}.

All baseline methods, alongside \ours{}, are implemented on \textbf{GPT-4V(ision)}~\cite{openai2023gpt4}, utilizing the \texttt{gpt-4-vision-preview} version to ensure a fair and consistent comparison.

\subsection{Implementation Details}
For geometry problem-solving and chess positional advantage prediction, we employ zero-shot prompting. In the case of molecular property prediction, we augment the instruction with four <SMILES, category> pairs, given the challenge this specialized task presents to the GPT-4V(ision). It is crucial to note that these SMILES representations are excluded from the test cases to prevent data leakage. Detailed instructions for these tasks can be found in Appendix~\ref{sec:appendix_instruction}.
To interact with the \texttt{gpt-4-vision-preview}, the \textit{temperature} and \textit{top\_p} are set to $0$ and $1$, respectively, to ensure deterministic outputs, while the \textit{max\_tokens} parameter is capped at $2048$.

\begin{table*}
\centering
\begin{tabular}{lcccccc}
\toprule
             & Level 1  & Level 2 & Level 3  & Level 4  & Level 5 & Avg.   \\
\midrule
\ours{}~(Ours)         & \textbf{71.43} & \textbf{53.13}   & \textbf{44.12} & 16.98 & \textbf{17.02} & \textbf{34.22} \\ \midrule
CoT$_m$     & 61.90 & 37.50  & 29.41   & \textbf{24.53} & 10.64 & 28.34 \\
CoT$_v$         & 52.38   & 37.50    & 26.47   & 13.21   & 10.64   & 23.53  \\
CoT$_d$          & 47.62   & 50.00  & 29.41  & 7.69    & 6.38    & 23.12  \\
\bottomrule
\end{tabular}
\caption{Evaluation results on the geometry problem-solving task. Numbers in bold indicate the best performance.}
\label{tab:exp_geo}
\end{table*}

\begin{table*}
\centering
\begin{tabular}{lcccc}
\toprule
             & Level 1  & Level 2 & Level 3  & Avg.   \\
\midrule
\ours{}~(Ours)          & \textbf{57.41} & \textbf{43.21}   & \textbf{41.67} & \textbf{46.99} \\ \midrule
CoT$_m$     & 51.85 & 37.04  & 39.58 & 42.08 \\
CoT$_v$         & 48.15   & 38.27    & 37.50   & 40.98  \\
CoT$_d$          & 46.30   & 33.33  & 39.58  & 38.80  \\
\bottomrule
\end{tabular}
\caption{Evaluation results on the chess positional advantage prediction task. Numbers in bold indicate the best performance.}
\label{tab:exp_chess}
\end{table*}

\subsection{Main Results}
\label{sec:exp_main}
The results of our experiments, presented in Table~\ref{tab:exp_main}, reveal several key observations:
(1) \ours{} surpasses all compared baseline methods, achieving relative improvements of 14.26\%, 10.25\%, and 6.30\% in geometry problem-solving, chess positional advantage prediction, and molecular property prediction, respectively. This superior performance can be attributed to \ours{}'s adeptness at leveraging the combined strengths of both visual and DSL representations, along with its capacity to pinpoint and address critical steps;
(2) The integration of DSL and visual information proves advantageous for multi-modal reasoning tasks. Our results demonstrate that CoT$_m$ achieves the second-best average performance, notably excelling in geometry problem-solving. This task benefits markedly from the complementary insights provided by DSL and visual inputs, indicating the value of integrating these modalities;
and (3) The process of effectively merging DSL representations with visual data poses a significant challenge, as evidenced by the subpar performance of CCoT. 

\section{Analysis}
\subsection{Ablation Study}
This ablation study evaluates four variants of our model across three datasets, as shown in Table~\ref{tab:exp_abl}. These variants comprise the full method and three variants: one without the diagnostic check~(``-diagnostic''), where the reasoning process is solely based on divergent reasoning chains from different modalities without any verification; one lacking image inputs~(``-visual''), where the model's assessment of reasoning chains relies exclusively on the DSL representation and its intrinsic knowledge; and one excluding DSL inputs~(``-dsl''), where the evaluation of reasoning chains depends solely on visual information and the model's inherent understanding. 

The results demonstrate that our full method outperforms all variants on the datasets, indicating the crucial role of combining DSL and visual inputs alongside diagnostic checks for identifying discrepancies and enhancing problem-solving in critical steps. Notably, the exclusion of visual inputs results in the most significant performance drop, highlighting the vital contribution of images to the efficacy of multi-modal reasoning tasks.

\subsection{Analysis on Different Complexities}
This experiment delves into how \ours{} performs under varying problem complexities, comparing it with three variants of chain-of-thought prompting.
Our focus is on geometry problem-solving and chess positional advantage prediction due to the labor-intensive nature of assessing the difficulty of molecular graphs. For geometry, we utilize the difficulty levels outlined by the MATH benchmark~\cite{hendrycks2021measuring}, and for chess, we classify problems into three difficulty levels based on the centipawns returned by Stockfish 15.

Table~\ref{tab:exp_geo} and Table~\ref{tab:exp_chess} present the results. \ours{} consistently outperforms competitors across nearly all difficulty levels, except level 4 in geometry problem-solving. Integrating DSL and image inputs proves advantageous, as CoT$_m$ typically surpasses the performance of both CoT$_v$ and CoT$_d$. However, achieving universal improvements through direct integration presents a significant challenge (as discussed in \S\ref{sec:pilot_versatile}). In geometry problem-solving, DSL representations are particularly effective in simpler problems, but this advantage diminishes with increased complexity. We hypothesize this is due to the lengthening of Asymptote code in more complex problems. For instance, the average Asymptote code length is $186.89$ for levels 1 to 3, but increases to $217.80$ for levels 4 to 5, whereas the length of FEN notation remains relatively stable across different levels of difficulty.

\begin{table*}
\centering
\begin{tabular}{lcccc}
\toprule
Methods & \textbf{G-MATH} & \textbf{ChessAdv} & \textbf{MUTAG}  & Avg. \\ \midrule
\ours{}~(Ours) & \textbf{34.22} & \textbf{46.99} & \textbf{83.52} & \textbf{54.71} \\ \midrule
Self-Refine~(2 turns) & 30.48 & 43.17 & 73.63 & 48.91 \\
Self-Refine~(3 turns) & 28.34 & 42.08 & 71.98 & 47.28 \\
Self-Refine~(4 turns) & 28.88 & 38.80 & 68.68 & 45.29 \\ 
\bottomrule
\end{tabular}
\caption{Comparative analysis of \ours{} versus Self-Refine prompting. Numbers in bold denote the best performance.}
\label{tab:exp_refine}
\end{table*}

\subsection{Comparison with Self-Refine Prompting}
\label{sec:exp_refine}
This experiment explores the efficacy of self-refine prompting~\cite{madaan2023self}, a technique that improves previous outputs through iterative feedback and refinement, as a potential substitute for the diagnostic check and alignment phases in \ours{}. We have adapted the conventional self-refine prompting approach to accommodate both DSL and image inputs, while preserving the original implementation details to the greatest extent. This experiment evaluates three versions of self-refine prompting, denoted as Self-Refine~($x$ turns), with $x-1$ indicating the count of refinement cycles and $x$ varying from $2$ to $4$. 

Table~\ref{tab:exp_refine} presents the results. The findings reveal that \ours{} consistently surpasses the various versions of self-refine prompting. This indicates the superiority of directing LVLMs to pinpoint inconsistencies between divergent solutions over merely generating feedback based on the knowledge embedded within their parameters. Moreover, recent work~\cite{huang2023large} corroborates our findings, demonstrating that LLMs frequently encounter difficulties in adjusting their responses based solely on their inherent capabilities.
This is further validated by our results, which indicate a decline in the performance of the self-refine prompting as the number of refinement iterations increases.

\subsection{Case Study}
Due to space constraints, the case study is included in Appendix~\ref{sec:appendix_case}.

\section{Related Work}

\subsection{Multi-Modal CoT Prompting}
An advanced methodology for zero-shot image reasoning leverages CoT prompting, a technique that breaks down complex tasks into simpler, sequential thought processes to simulate human reasoning~\cite{lu2022learn,zhang2023multimodal,wang2023filling}. Due to the structural differences between LVLMs and LLMs, additional improvements have been made to adapt CoT for wider applications. To illustrate, QVix~\cite{yang2023good} leverages LLMs' linguistic skills to enhance LVLMs' visual content analysis; V$^*$~\cite{wu2023textit} enhances the precise targeting of specific visual elements; \citet{wu2023role} address CoT prompting's limitations by adopting a ``Description then Decision'' strategy for complex visiolinguistic tasks; CoCoT~\cite{zhang2024cocot} uses a contrastive CoT approach for multiple image inputs; ViLa~\cite{hu2023look} merges perceptual data with CoT for physically-grounded task planning; and DDCoT~\cite{zheng2023ddcot} assigns tasks to relevant components, differentiating reasoning and recognition roles and integrating visual recognition into the reasoning process. Despite these advancements, the strategic use of prompting mechanisms to seamlessly integrate DSLs into LVLMs presents an untapped potential, a gap our research aims to bridge by pioneering in this specific area.

\subsection{Multiple Chains Prompting}
\label{sec:related_multi}
Following the progress of the chain-of-thought prompting, a series of efforts have been made to enhance factuality by generating multiple reasoning chains. Building on this progress, the research focuses on three main approaches: self-consistency~\cite{wang2023selfconsistency}, self-refinement~\cite{madaan2023self,shinn2023reflexion,chen2023teaching}, and multi-agent debate~\cite{du2023improving,liang2023encouraging,xiong2023examining}. Self-consistency~\cite{wang2022self} involves a method where various reasoning paths are first generated, and then the most consistent answer is selected through a process akin to majority voting. Self-refinement~\cite{madaan2023self} leverages the inherent capabilities of LLMs to generate feedback for previous outputs, refining them based on this feedback. However, recent research~\cite{huang2023large} indicates that LLMs face challenges in providing accurate feedback independently, suggesting that feedback from external environments~\cite{first2023baldur} is a more effective alternative. Multi-agent debate~\cite{du2023improving} aims to replicate real-world debate scenarios, fostering a consensus by incorporating outputs from previous iterations in each debate cycle. These methods, while innovative, have yet to fully address the need for identifying intermediate inconsistencies between multiple chains which play a crucial role in pinpointing the critical steps necessary for solving complex tasks. Moreover, the requirement for multiple invocations of LLMs, particularly with proprietary LVLMs~\cite{openai2023gpt4}, significantly increases the associated costs.

We provide a detailed review of the literature on large vision-language models in Appendix~\ref{sec:appendix_related}.

\section{Conclusion}
In conclusion, our work introduces the Bi-Modal Behavioral Alignment (\ours{}) prompting method, a novel approach that significantly enhances the multimodal reasoning capabilities of GPT-4V(ision) by integrating DSL. By generating and aligning separate reasoning chains for visual and DSL representations, \ours{} addresses the challenges of inconsistent reasoning mechanisms and the execution of multi-step reasoning tasks. Our experiments across diverse domains, including geometry problem-solving, chess positional advantage prediction, and molecular property prediction, demonstrate the effectiveness of \ours{}, showcasing notable improvements in performance.

\section*{Ethical Considerations}
In adherence to the established Code of Ethics, this work exclusively employs publicly accessible data and information, ensuring no private or confidential resources are utilized.

\section*{Limitations}
\ours{} marks a significant advancement in the field of multi-modal reasoning, incorporating DSLs. Despite this, it is beneficial to address several limitations to fully exploit its capabilities:

(1) \ours{} demonstrates significant improvements in three distinct domains: geometry, chess, and molecular biology. Yet, its application in other areas, especially those without custom DSLs, has not been extensively explored. Adapting \ours{} by substituting DSL representations with alternative, advanced representations, such as scene graphs~\cite{yang2018graph}, could be advantageous. These alternatives, although less precise and informative in capturing image nuances, offer a valuable research direction.

(2) The primary aim of this work is to develop a prompting method, that complements, but is distinct from, other advanced technologies~\cite{yao2022react,xie2023openagents}. The possibility of integrating and responding to environmental feedback to develop a more adaptive and intelligent agent is an intriguing future research direction.

\bibliography{anthology,custom}
\bibliographystyle{acl_natbib}

\clearpage
\appendix

\section{More Details about Pilot study}

\begin{figure*}[t]
    \centering
    \includegraphics[width=0.9\textwidth]{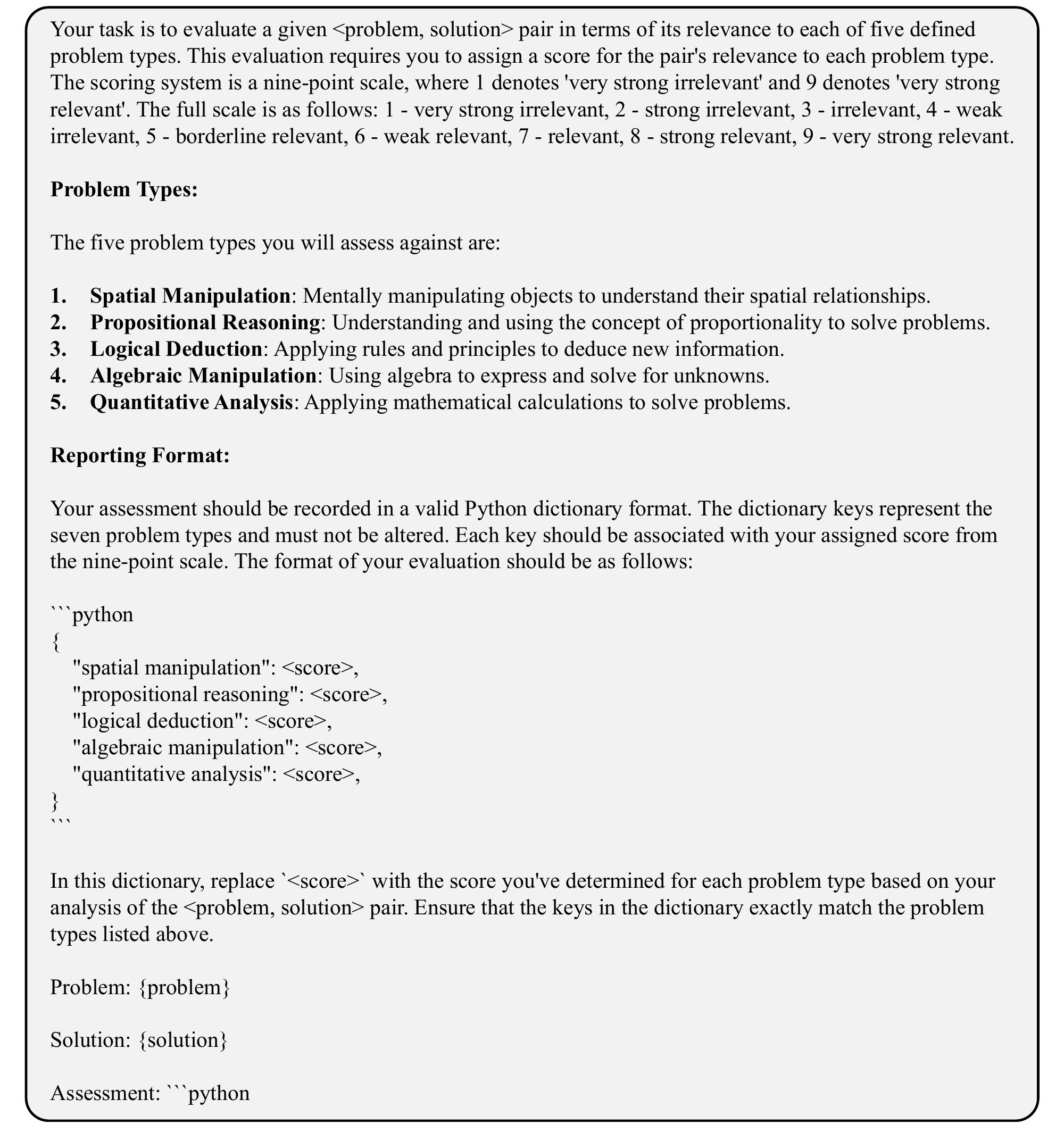}
    \caption{Illustration of the prompt utilized for category annotation.}
    \label{fig:pilot_prompt_1}
\end{figure*}

\subsection{Annotation of Problem Types}
\label{sec:appendix_type}
In this experiment, we (the authors) identified five primary categories essential for problem-solving in geometry: Spatial Manipulation, Propositional Reasoning, Logical Deduction, Algebraic Manipulation, and Quantitative Analysis. These categories were conceptualized based on our extensive experience in geometry problem-solving, where Spatial Manipulation and Propositional Reasoning are aligned with intuitive reasoning, and the remaining categories, namely Logical Deduction, Algebraic Manipulation, and Quantitative Analysis are associated with precise, calculation-based approaches.

For the categorization of problems, we employed \texttt{gpt-4-1106-preview} to label each problem according to these predefined categories. This process was facilitated using a prompt, as detailed in Figure~\ref{fig:pilot_prompt_1}, wherein placeholders for the problem statement and its solution were substituted with the actual content. The categorization was determined based on the highest-scoring category for each problem. In instances where multiple categories achieved the highest score, the final classification was randomly assigned from among these top-scoring categories.

\begin{figure*}[t]
    \centering
    \includegraphics[width=0.9\textwidth]{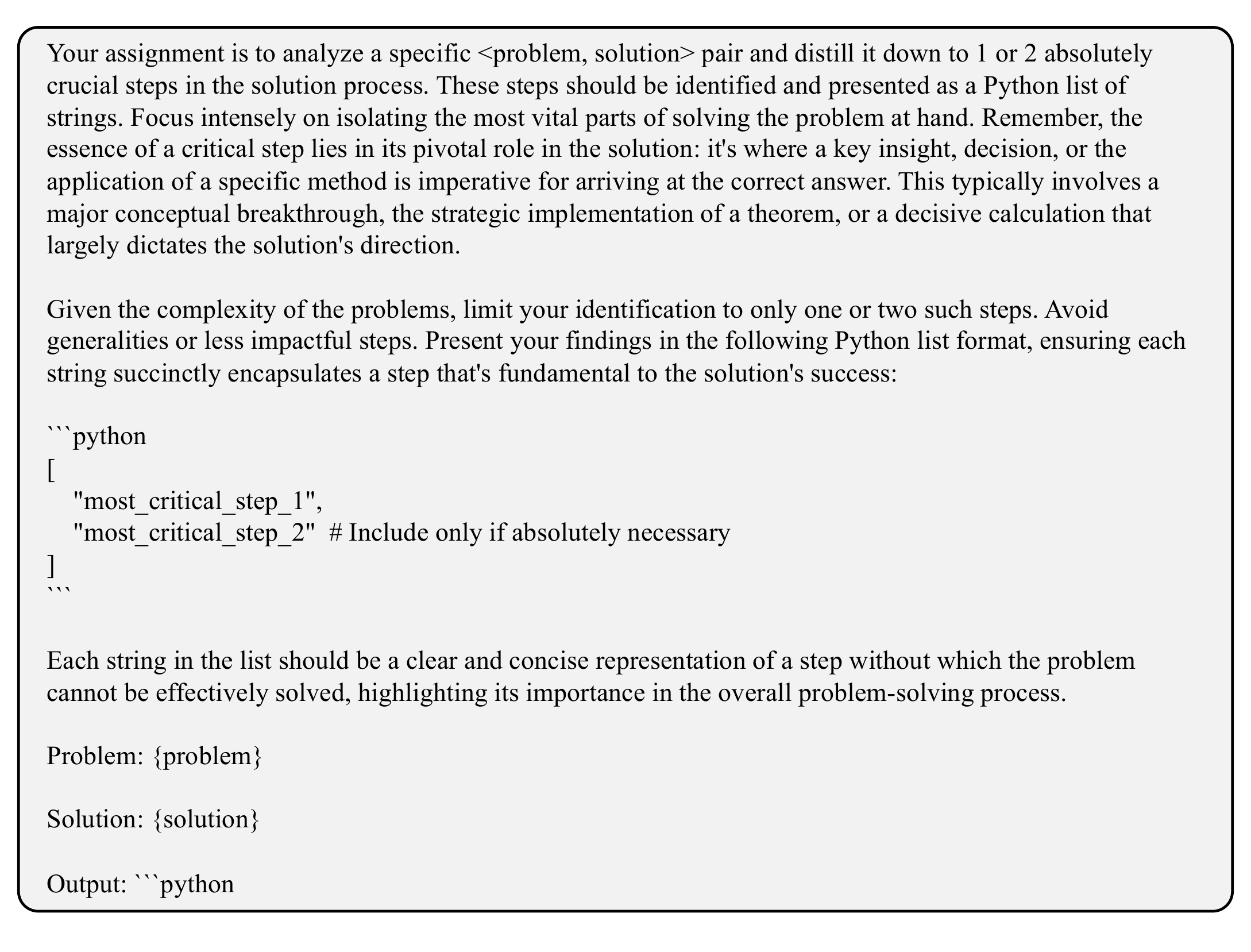}
    \caption{Illustration of the prompt utilized for critical step identification.}
    \label{fig:pilot_prompt_2_1}
\end{figure*}

\begin{figure*}[t]
    \centering
    \includegraphics[width=0.9\textwidth]{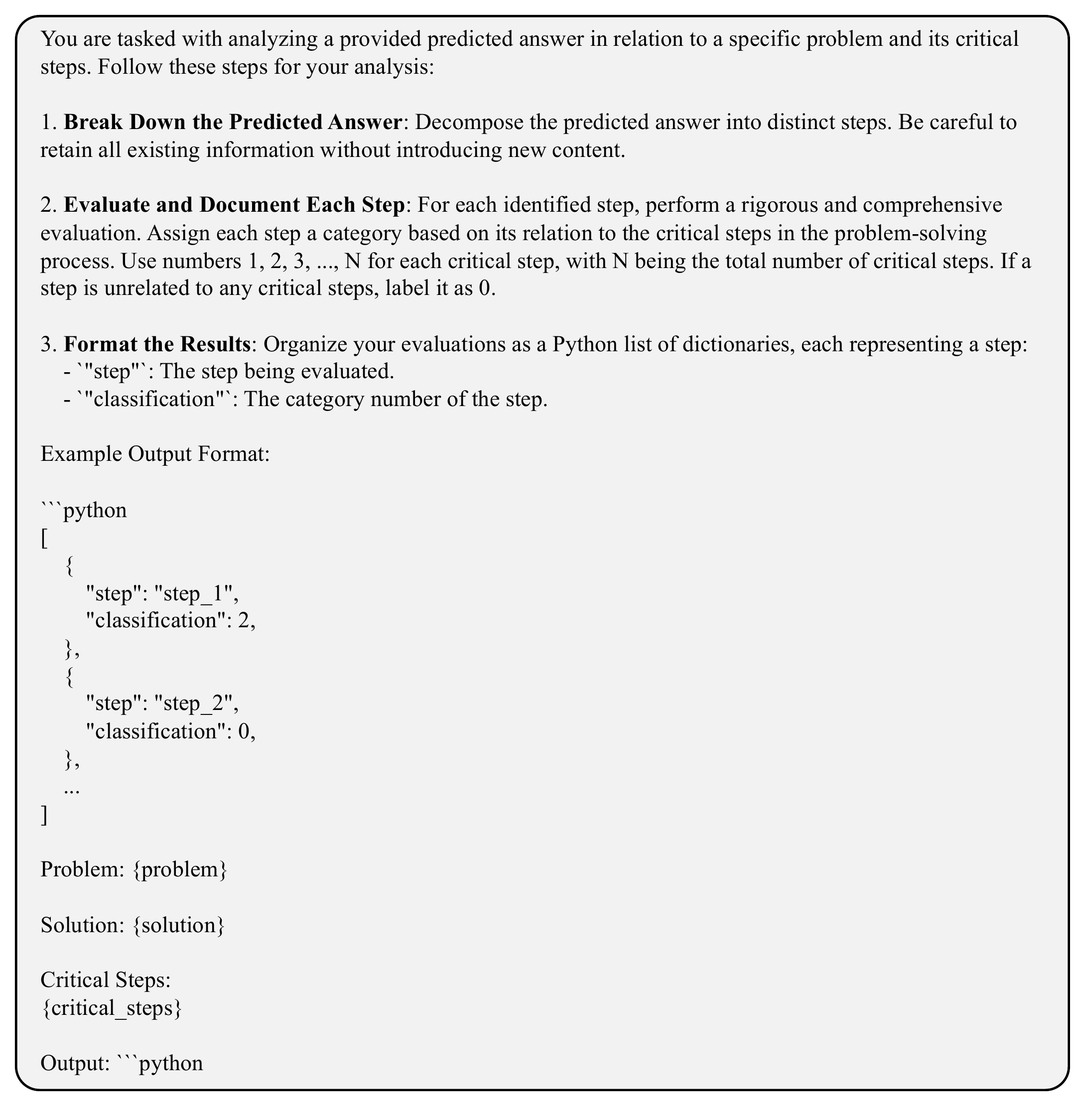}
    \caption{Illustration of the prompt utilized for categorizing each step within the generated solution.}
    \label{fig:pilot_prompt_2_2}
\end{figure*}

\subsection{Annotation of Critical Steps}
\label{sec:appendix_critical}
The prompts employed for the annotation of the critical steps associated with each problem are depicted in Figures \ref{fig:pilot_prompt_2_1} and \ref{fig:pilot_prompt_2_2}. Specifically, Figure \ref{fig:pilot_prompt_2_1} illustrates the instructions utilized to identify the critical steps necessary for solving the problem, alongside the provision of a ground-truth solution. This is necessitated by the current limitations of LLMs in addressing geometry problems. Figure \ref{fig:pilot_prompt_2_1} demonstrates the instructions employed for categorizing each step within the generated solution as either corresponding to one of the previously identified critical steps or not. All annotations were performed using \texttt{gpt-4-1106-preview} The accuracy of these critical annotations was subsequently verified by the authors.

\section{System Instructions}
\label{sec:appendix_instruction}
\begin{figure*}[t]
    \centering
    \includegraphics[width=0.9\textwidth]{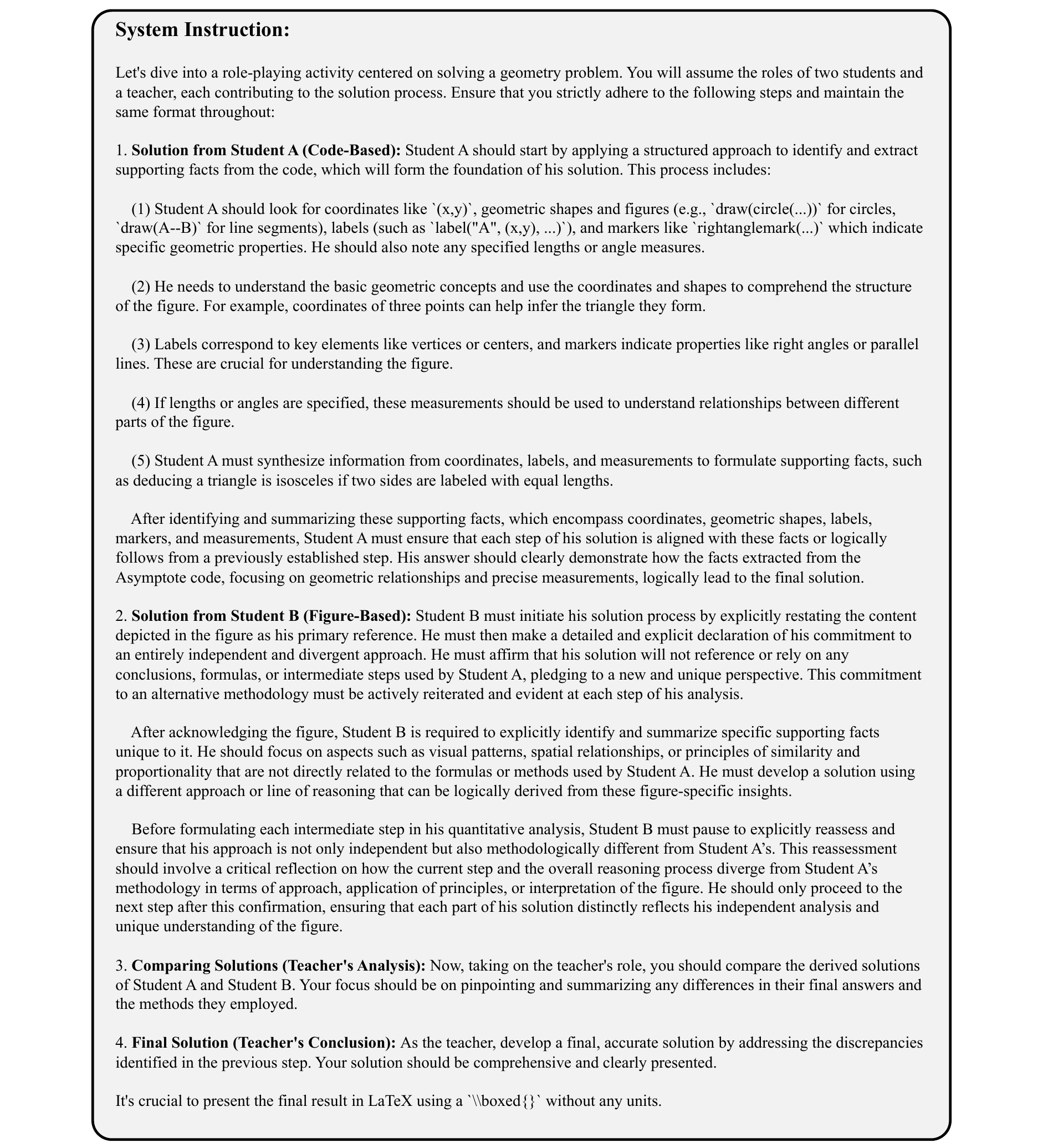}
    \caption{System instruction for geometry problem-solving.}
    \label{fig:system_geo}
\end{figure*}

\begin{figure*}[t]
    \centering
    \includegraphics[width=0.9\textwidth]{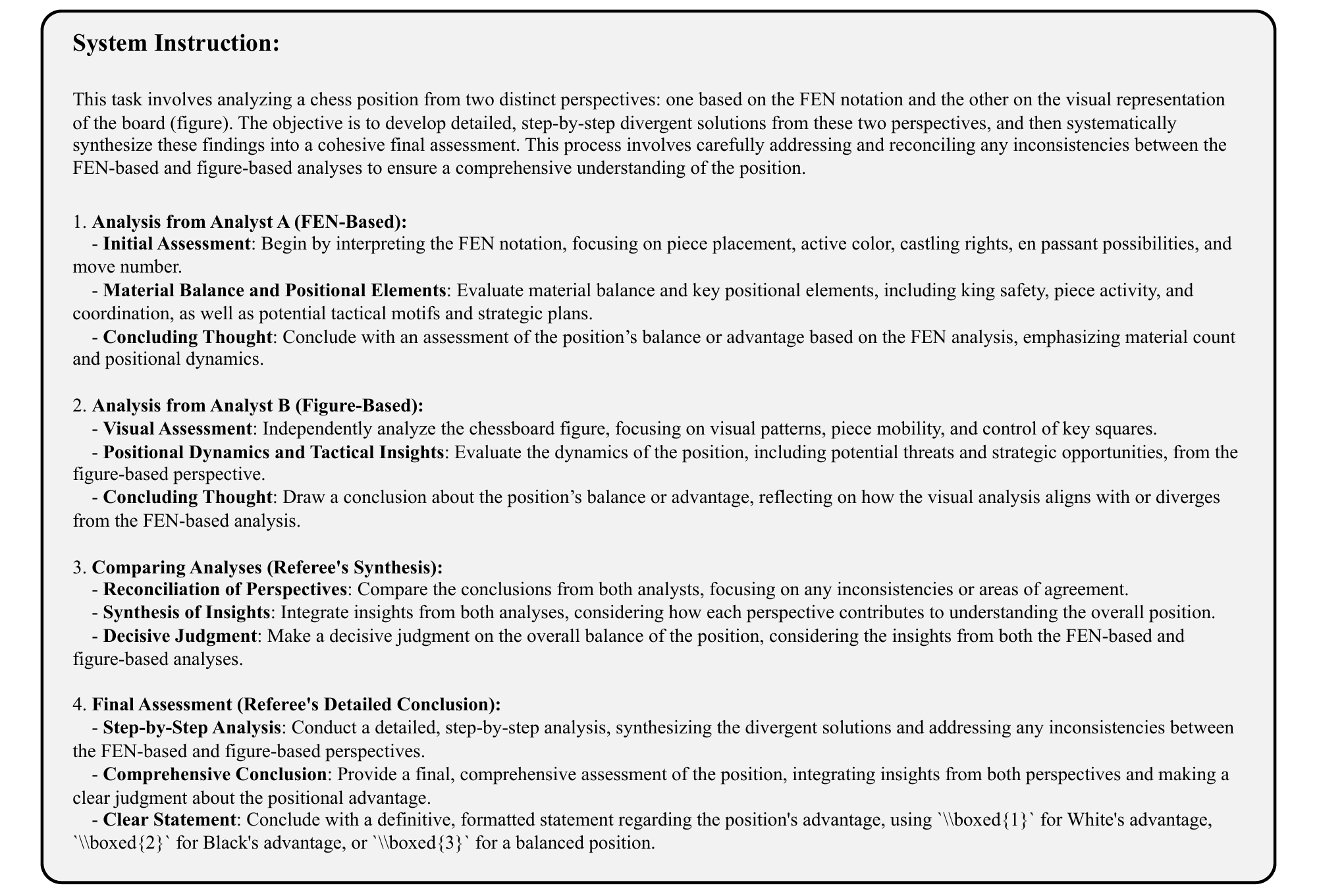}
    \caption{System instruction for chess positional advantage prediction.}
    \label{fig:system_chess}
\end{figure*}

\begin{figure*}[t]
    \centering
    \includegraphics[width=0.9\textwidth]{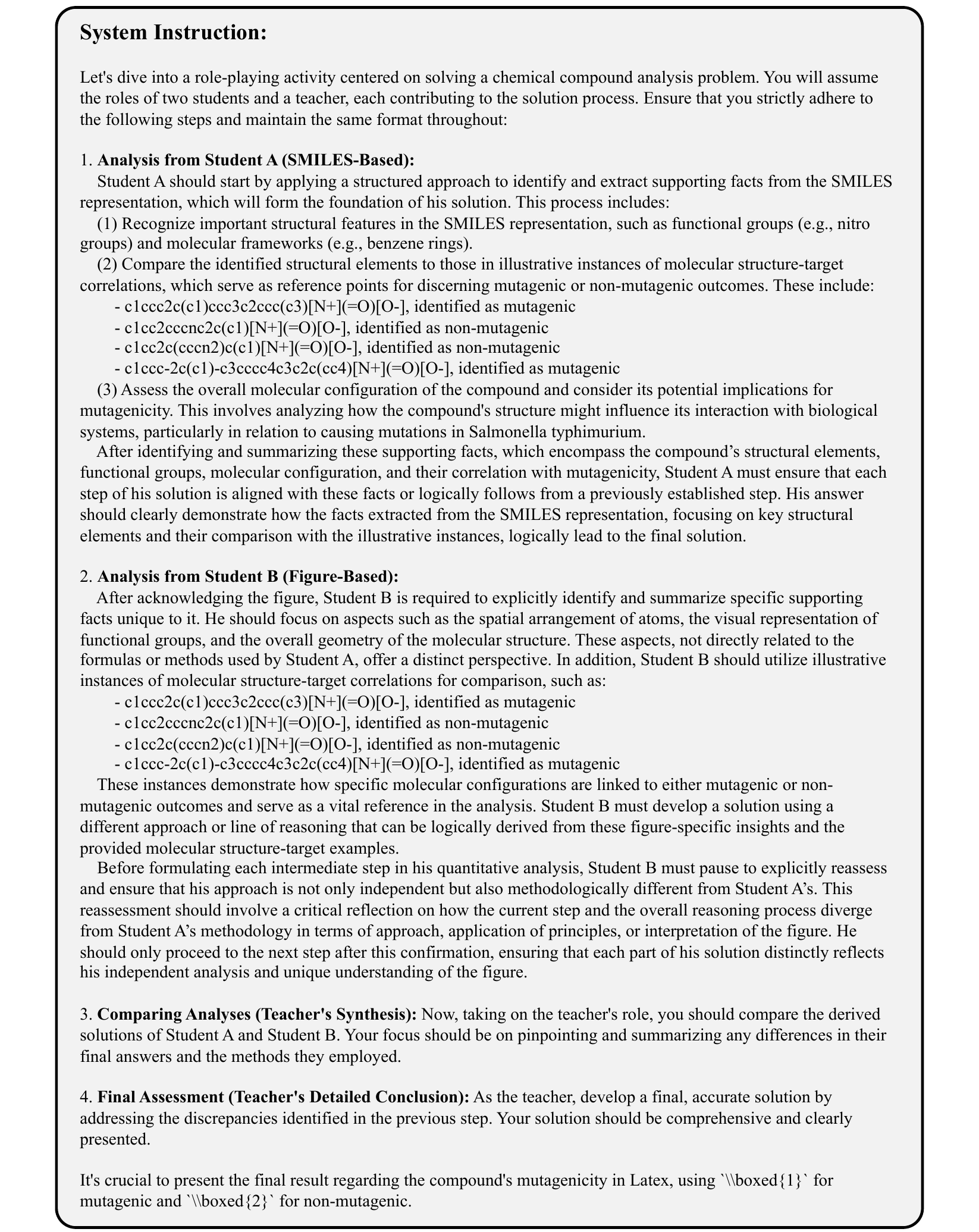}
    \caption{System instruction for molecular property prediction.}
    \label{fig:system_molecular}
\end{figure*}

In this work, we employ system instructions~\cite{mukherjee2023orca} to guide LVLMs toward producing the requisite outputs across various stages, as elaborated in \S\ref{sec:method}. The specific system instructions applied for geometry problem-solving, chess positional advantage prediction, and molecular property prediction are illustrated in Figures \ref{fig:system_geo}-\ref{fig:system_molecular}.

\section{More Discussions about Related Work}
\label{sec:appendix_related}
\subsection{Large Vision-Language Models}
Driven by the success of LLMs, research in vision-language models has begun to explore how to equip LLMs with the ability to process visual inputs to solve a variety of multimodal tasks~\cite{alayrac2022flamingo,liu2023visual,zhu2023minigpt,peng2023kosmos}. These advancements leverage the extensive prior knowledge and intricate reasoning capabilities established by recent LLMs~\cite{touvron2023llama,touvron2023llama2}, integrating them with visual modalities to enhance performance across diverse applications. A common approach in this field involves using a vision model to transform images into visual features, which are then aligned with the LLMs' feature space through a bridge module. This alignment is achieved either by employing a linear or MLP layer as a bridge module~\cite{liu2023improved,liu2023visual,chen2023shikra,gao2023llama,peng2023kosmos,zhang2023llama,wang2023cogvlm} or by designing more complex bridge networks to compress or adaptively select visual information~\cite{alayrac2022flamingo,zhu2023minigpt,li2023fine,dai2023instructblip,li2023blip,awadalla2023openflamingo,zhao2023mmicl,li2023otter,ye2023mplug,bai2023qwen}. The datasets employed to train these bridge modules are notably varied, ranging from synthetic image-text pairs~\cite{li2023fine} and large-scale multi-modal web corpora~\cite{alayrac2022flamingo} to specialized datasets for conversational~\cite{liu2023visual} and academic purposes~\cite{liu2023improved}.
Notably, the emergence of advanced, proprietary LVLMs such as GPT-4V(ision)~\cite{openai2023gpt4} and Gemini~\cite{team2023gemini} has set new benchmarks by demonstrating superior performance in multimodal task execution and cognitive reasoning, surpassing human expertise.
While these models have achieved significant success in vision-language tasks, the exploration of unlocking the full potential of DSLs, especially for intricate multi-modal reasoning tasks, remains largely unexplored.

\section{Case Study}
\label{sec:appendix_case}
\begin{figure*}[t]
    \centering
    \includegraphics[width=1.0\textwidth]{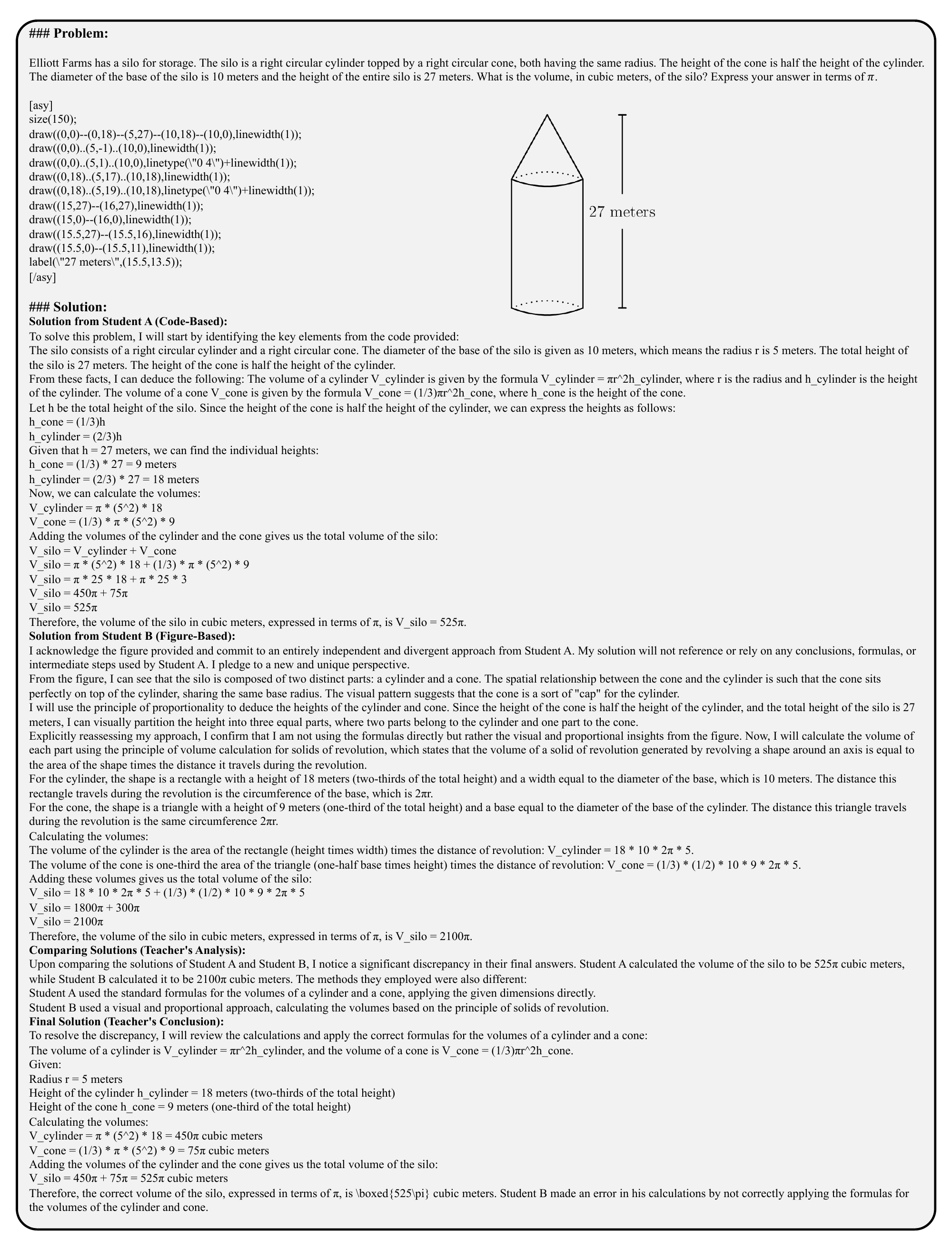}
    \caption{Case study 1: geometry problem-solving.}
    \label{fig:case_geo_1}
\end{figure*}

\begin{figure*}[t]
    \centering
    \includegraphics[width=1.0\textwidth]{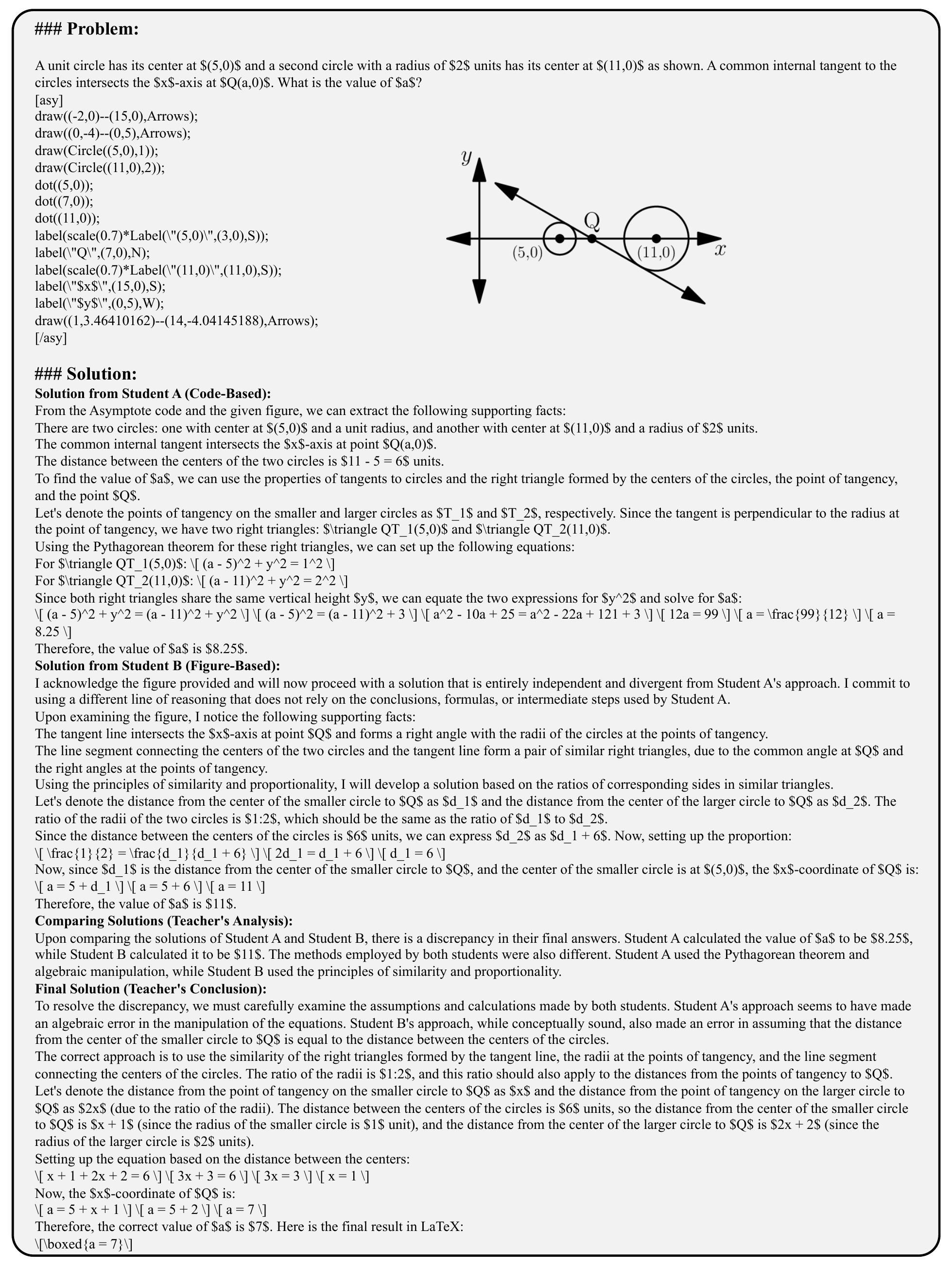}
    \caption{Case study 2: geometry problem-solving.}
    \label{fig:case_geo_2}
\end{figure*}

\begin{figure*}[t]
    \centering
    \includegraphics[width=1.0\textwidth]{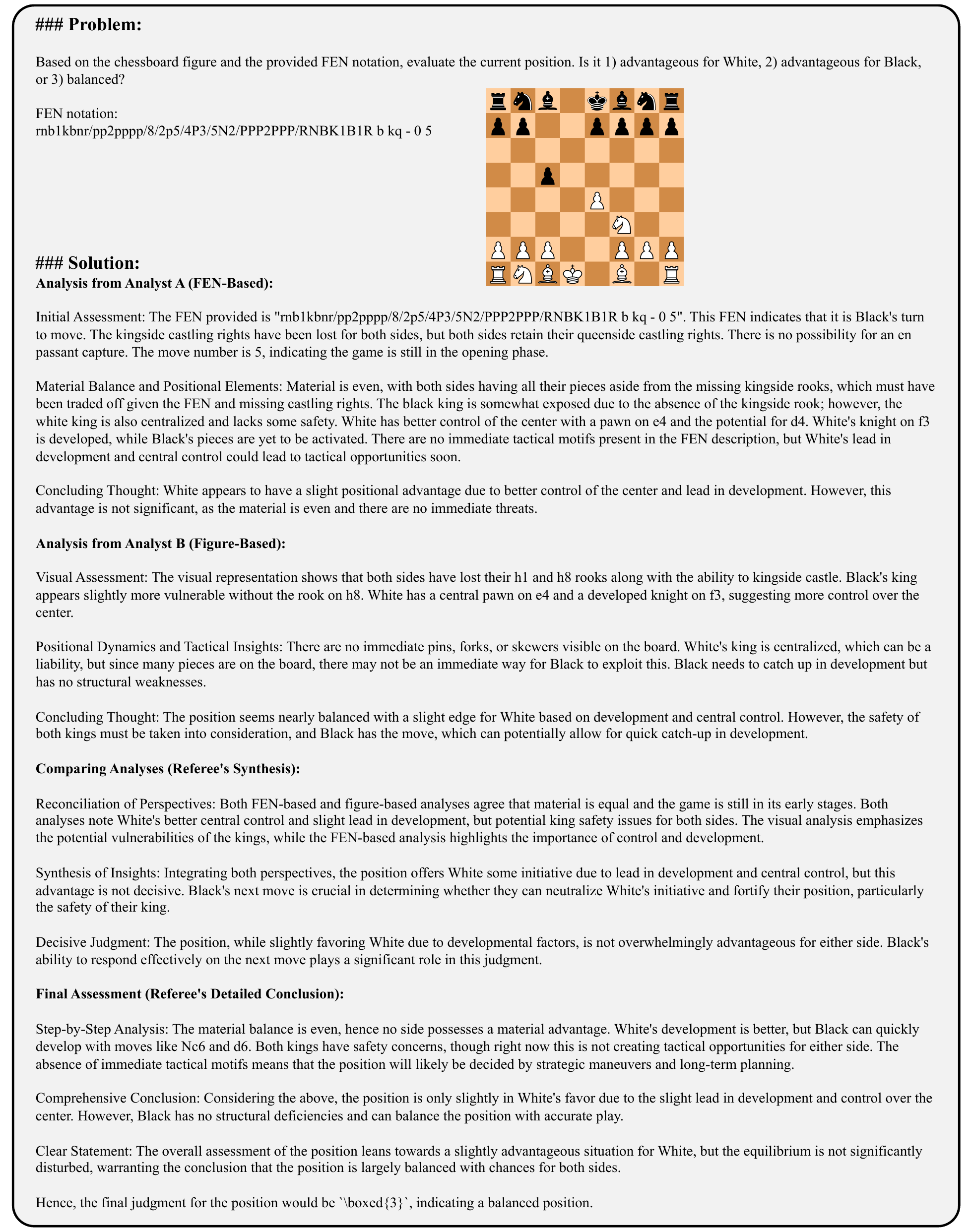}
    \caption{Case study 3: chess positional advantage prediction.}
    \label{fig:case_chess_1}
\end{figure*}

\begin{figure*}[t]
    \centering
    \includegraphics[width=1.0\textwidth]{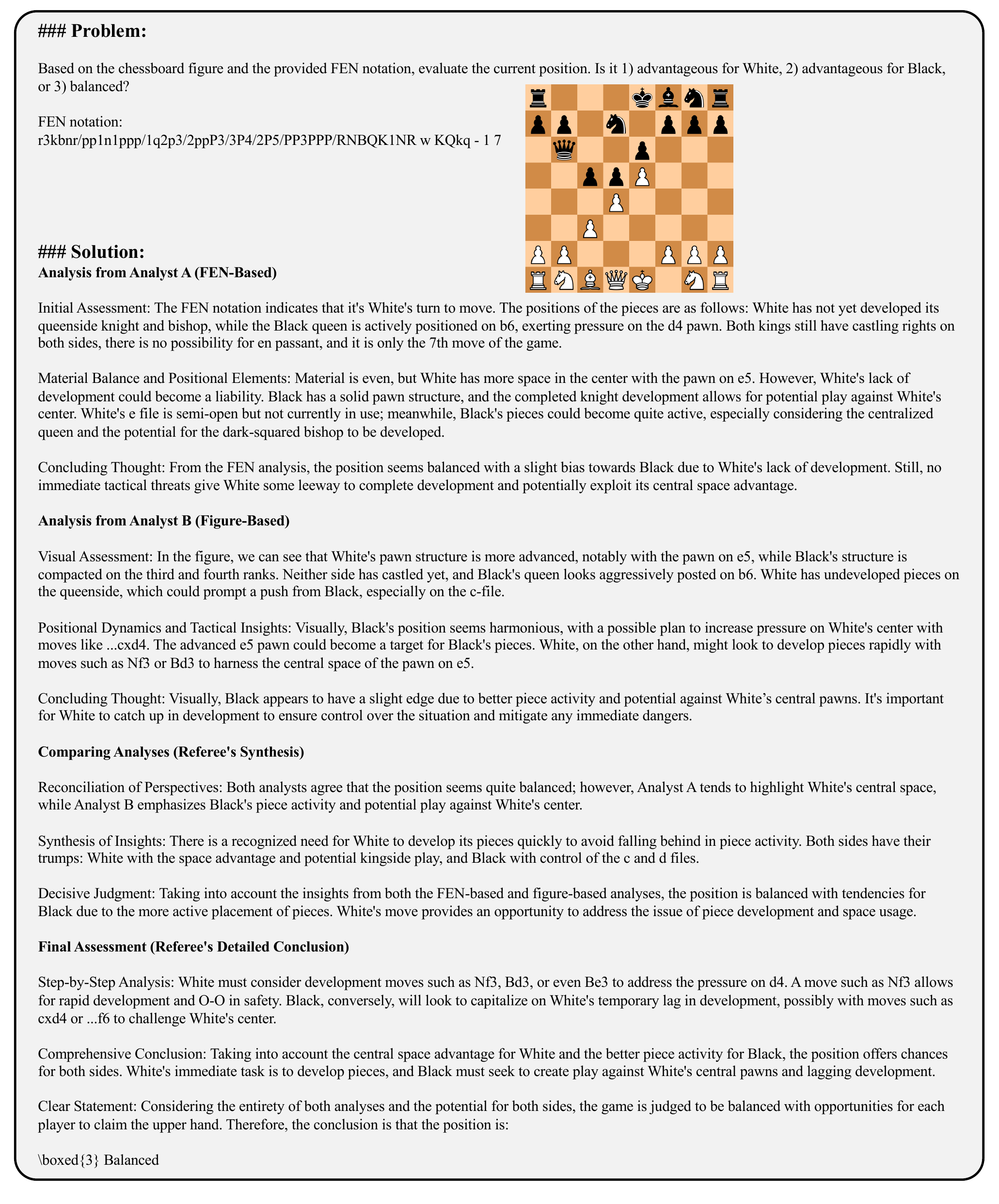}
    \caption{Case study 4: chess positional advantage prediction.}
    \label{fig:case_chess_2}
\end{figure*}

\begin{figure*}[t]
    \centering
    \includegraphics[width=1.0\textwidth]{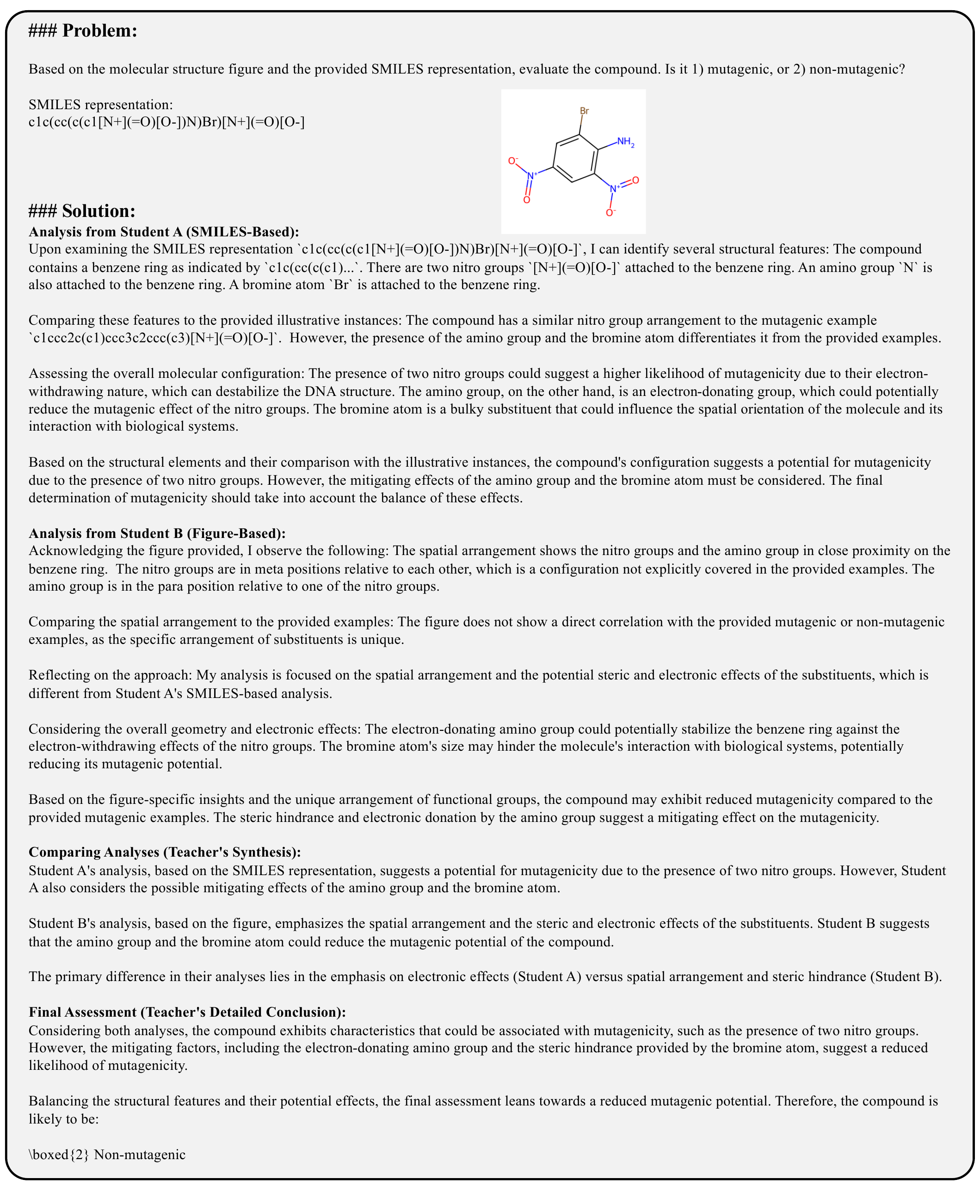}
    \caption{Case study 5: molecular property prediction.}
    \label{fig:case_molecular_1}
\end{figure*}

\begin{figure*}[t]
    \centering
    \includegraphics[width=1.0\textwidth]{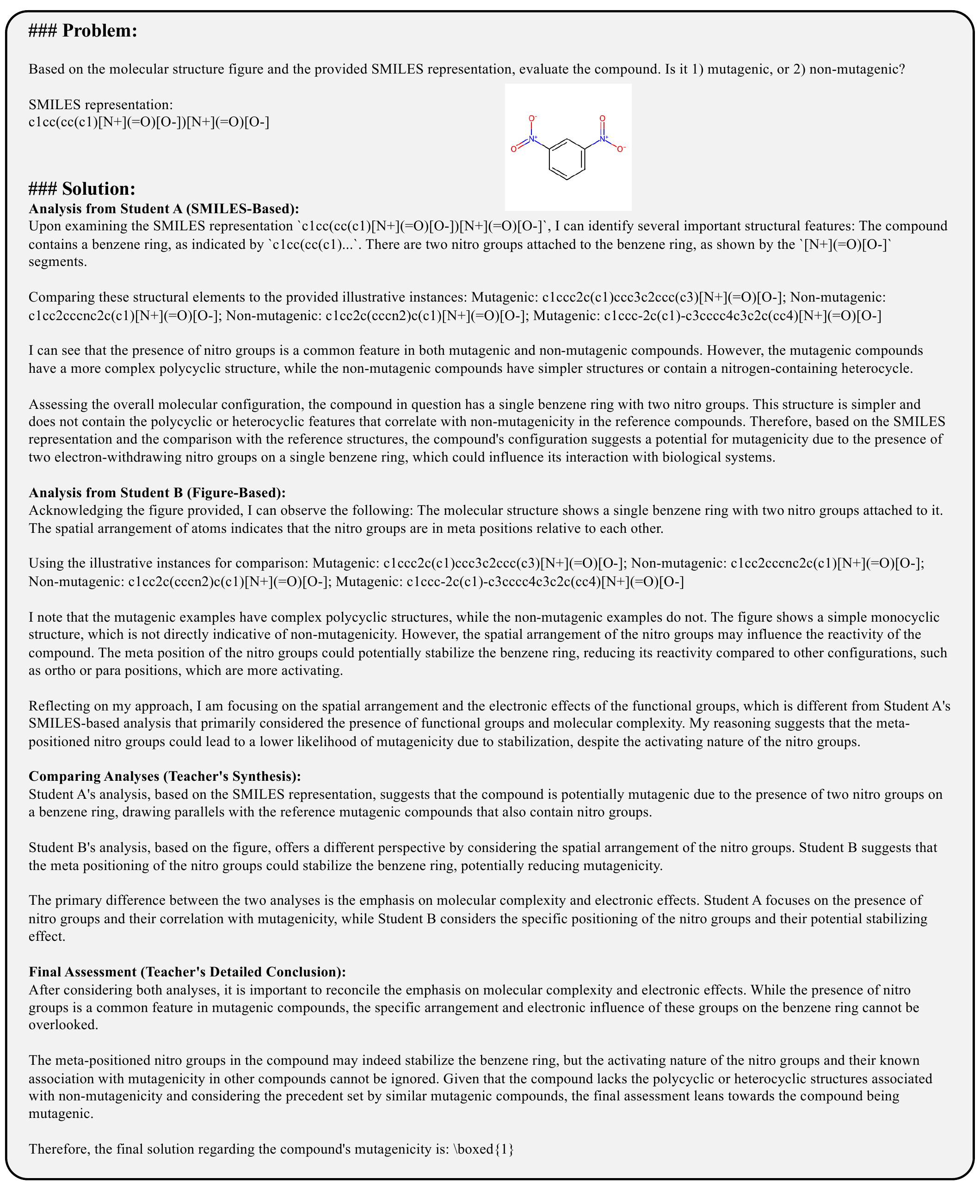}
    \caption{Case study 6: molecular property prediction.}
    \label{fig:case_molecular_2}
\end{figure*}

To enhance our understanding of the effectiveness of \ours{}, we showcase a series of outputs in Figures \ref{fig:case_geo_1}-\ref{fig:case_molecular_2}. These illustrations demonstrate that our approach is capable of accurately identifying discrepancies between two chains. This identification process is subsequently leveraged to synthesize the final solution.

\end{document}